\documentclass[10pt,twocolumn,letterpaper]{article}

\usepackage[pagenumbers]{arxiv_twocolumn} 

\usepackage{url}
\usepackage{dsfont}
\usepackage{afterpage}

\usepackage{booktabs} 
\usepackage{diagbox}  
\usepackage{array}    
\usepackage{arydshln}
\usepackage{multirow} 
\usepackage{colortbl} 
\usepackage{bbm}
\usepackage{algorithm}
\usepackage{algpseudocode}                  
\usepackage{setspace}                       
\usepackage{lipsum} 

\usepackage{wrapfig}

\usepackage{mathtools} 
\usepackage{bm} 
\usepackage{soul} 
\usepackage{nicefrac}
\usepackage{csquotes}

\usepackage{duckuments}

\usepackage{pifont} 
\makeatletter
\newcommand*{\myfnsymbol}[1]{%
  \ensuremath{%
    \ifcase#1\or
      \text{\ding{41}}\or
      1\or
      2\or
      3\or
      4\or
      5\or
      6\or
      7\else
      8\fi
  }%
}
\def\@fnsymbol#1{\myfnsymbol{#1}}
\def\thempfootnote{\myfnsymbol{\c@mpfootnote}}
\makeatother

\usepackage[most]{tcolorbox}
\usepackage{enumitem}

\newcommand{\takeaway}[2]{%
    \begin{tcolorbox}[
        enhanced,
        colback=gray!4,
        colframe=cyan!80!black, 
        boxrule=0.8pt,
        arc=4mm,
        left=3mm,
        right=3mm,
        top=2mm,
        bottom=2mm,
        before skip=1em,
        after skip=1em,
    ]
    \begin{itemize}[leftmargin=1.25em, nosep]
        \item \textbf{Takeaway #1.} \emph{#2}
    \end{itemize}
    \end{tcolorbox}%
}

\definecolor{cvprblue}{rgb}{0.21,0.49,0.74}
\usepackage[pagebackref,breaklinks,colorlinks,allcolors=cvprblue]{hyperref}

\def\paperID{*****} 
\def\confName{CVPR}
\def\confYear{2026}

\title{An Empirical Study of Training Pixel-Space\\
 Text-to-Image  Diffusion Models}

\author{
\vspace{-2em}\\
\fontsize{10.4pt}{9.84pt}\selectfont
\textbf{Dengyang Jiang}$^{2,1}$ ~
\textbf{Ruoyi Du}$^{1}$ ~
\textbf{Zhennan Chen}$^{3}$ ~
\textbf{Dongyang Liu}$^{1}$ ~ \\
\fontsize{10.4pt}{9.84pt}\selectfont
\textbf{Zanyi Wang}$^{4}$ ~
\textbf{Mingzhe Zheng}$^{2}$ ~
\textbf{Xiangpeng Yang}$^{1}$ ~
\textbf{Huanqia Cai}$^{1}$ ~ \\
\fontsize{10.4pt}{9.84pt}\selectfont
\textbf{Aiming Hao}$^{1}$ ~ 
\textbf{Yuming Jiang}$^{1}$ ~
\textbf{Peng Gao}$^{1}$\thanks{Corresponding authors} ~ \hspace{1mm}
\textbf{Harry Yang}$^{2}\textsuperscript{\ding{41}}$ ~
\textbf{Steven Hoi}$^{1}$ 
\\[1.5mm]
\fontsize{10pt}{9.5pt}\selectfont
$^{1}$Alibaba Token Hub, Alibaba Group ~
$^2$The Hong Kong University of Science and Technology \\
\fontsize{10pt}{9.5pt}\selectfont
$^{3}$Nanjing University ~~
$^{4}$University of California, San Diego 
}

\makeatletter
\g@addto@macro\@maketitle{
\vspace{-2em}
\begin{figure}[H]
\setlength{\linewidth}{\textwidth}
\setlength{\hsize}{\textwidth}
\centering
\includegraphics[width=\textwidth]{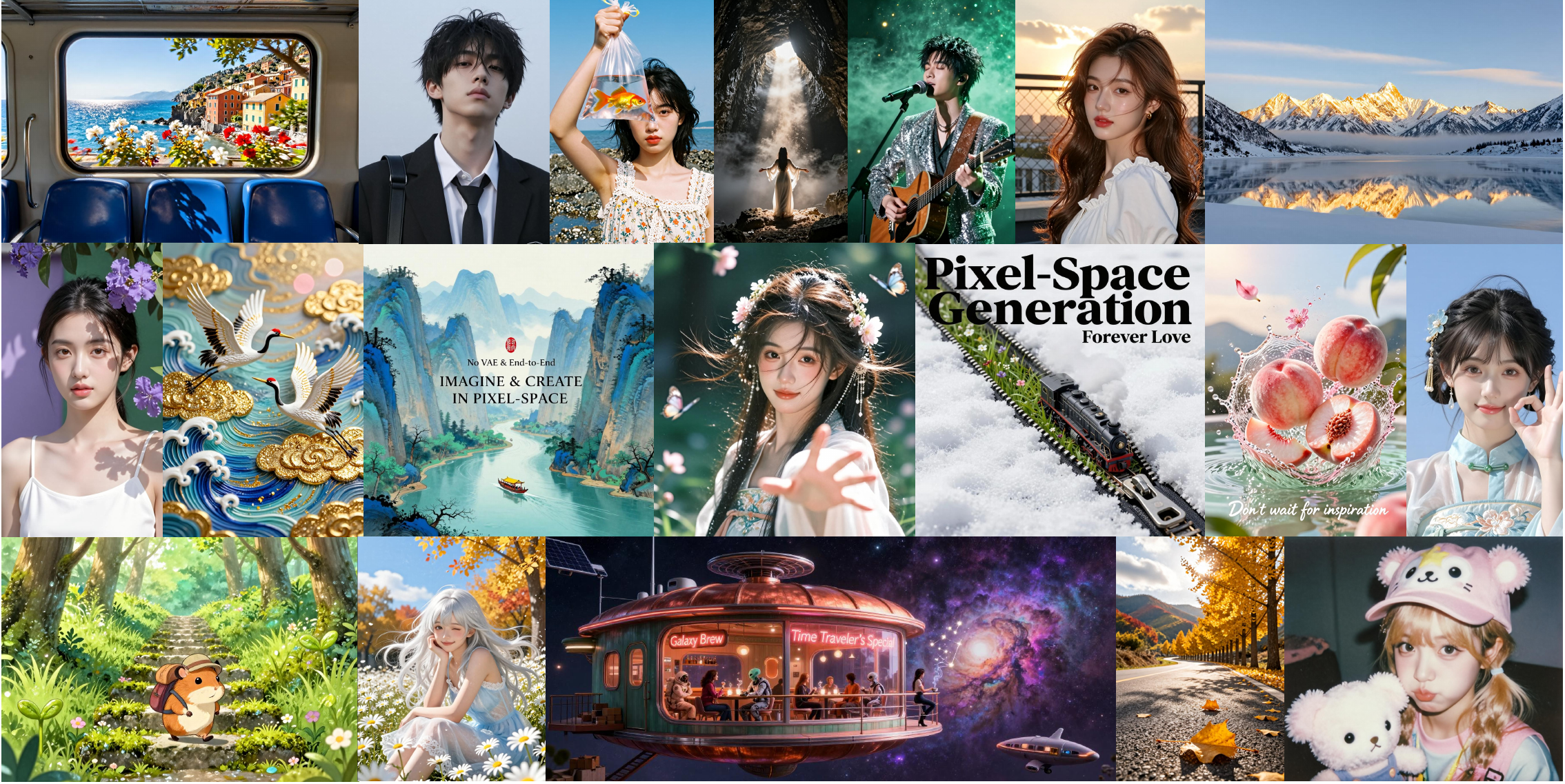}
\caption{\textbf{Images generated by Z-Image-Turbo (Pixel) trained through our method.} Offering only $\sim$ 0.2 second inference latency on an enterprise-grade H800 GPU.}

\label{fig:begin_vis}
\end{figure}
}
\makeatother

\begin{document}
\maketitle

\begin{abstract}
This paper investigates an increasingly important topic in generative modeling: pixel-space diffusion models. Although numerous studies have explored this topic, most focus on small-scale or class-conditional settings. Consequently, a practical recipe for training pixel-space models that rival or exceed well-established latent-space counterparts remains elusive. Through a comprehensive empirical study, we first observe that direct large-scale pre-training in pixel space converges substantially more slowly than in latent space. This observation motivates a latent-to-pixel strategy that acquires generative priors efficiently in latent space and transitions to pixel space during post-training. We then systematically investigate the key design choices governing this transition, including weight initialization, data composition, prediction target, decoder architecture, and noise schedule, and identify a practical recipe that makes the resulting pixel-space models match or outperform their latent-space counterparts while delivering 3.18 to 4.75 times end-to-end inference speedups. We hope that our findings provide useful empirical insights and practical guidelines for future research on pixel-space generation.
\end{abstract}


\section{Introduction}
\label{sec:intro}
Pixel-space diffusion has emerged as a  prominent paradigm, attracting substantial interest in the community~\cite{jit,pixnerd,pixeldit,dip,pixelflow,minit2i,latent-forcing}. In comparison with the dominant latent-space counterpart, it exhibits several distinct advantages: (i) by learning directly from RGB images and generating outputs in pixel space, it is not constrained by the reconstruction ceiling of a pre-trained VAE and can recover visual information discarded by latent compression~\cite{va-vae,pixeldit,ifid,sd-vae}. (ii) it offers practical efficiency benefits at inference time: images are generated directly as pixels without an additional VAE decoding stage, while the sequence length can be reduced by adopting a larger patch size. This provides a flexible efficiency - quality trade-off without increasing the VAE compression ratio, which often causes substantial degradation in reconstruction and generation quality~\cite{qwen-image-vae,ltx-video,dcae,zitpp,pixverve,l2p}.

Although the training recipes and scaling behavior of the latent-space diffusion have been extensively studied at scale through sustained community effort~\cite{sd3,dit,sit,seedream2,qwenimage,z-image,pixartalpha,dalle}, pixel-space diffusion remains far less explored. Most existing studies are still confined to class-conditioned ImageNet~\cite{jit,dip,pixnerd,latent-forcing} or small-scale text-to-image synthesis~\cite{pixelgen,pixeldit,deco,minit2i} with limited data~\cite{blip3o,share-gpt-4o-image,echo-4o}. Therefore, it remains largely unclear how to train a pixel-space counterpart with capabilities comparable to well-established latent-space text-to-image diffusion models~\cite{hunyuanimage,qwen-image-2,z-image,flux-2}. In particular, two fundamental questions remain unanswered: how do pixel- and latent-space diffusion compare in training efficiency under a large-scale training settings, and what training recipe can make pixel-space models competitive in both generation quality and inference efficiency?

In this work, we revisit the foundations of pixel-space model training and  investigate how to build an efficient and competitive pixel-space text-to-image model at scale. We first show that, under identical large-scale pre-training settings, pixel-space diffusion converges substantially more slowly than its latent-space counterpart. This motivates a latent-to-pixel strategy~\cite{l2p,asymmetric-flow} that leverages latent-space pre-training for efficient knowledge acquisition and introduces pixel-space learning during post-training. We then study the key design choices governing this transition, including initialization, training data, prediction target, decoder architecture, and noise schedule.
Based on these findings, we further combine progressive patch-size adaptation with step distillation to improve inference efficiency. Our final pixel-space models maintain competitive overall benchmark performance while delivering $3.18\times$--$4.75\times$ end-to-end speedups over their latent-space counterparts. Compared with prior latent-to-pixel methods~\cite{l2p,asymmetric-flow}, our models improve most benchmarks while substantially reducing inference latency. Consistent results on both Z-Image and FLUX2-klein demonstrate that the resulting recipe generalizes beyond a single model family.

In summary, our main contributions are as follows:
\vspace{-0.07in}
\begin{itemize}[leftmargin=*]
    \item We provide a controlled large-scale comparison of pixel- and latent-space diffusion, revealing the slower convergence of pixel-space pre-training.
    \item We systematically study latent-to-pixel adaptation at scale and establish a practical recipe for effective pixel-space post-training.
    \item Our final recipe produces pixel-space models that achieve comparable overall performance to the well-established latent-space counterparts while providing $3.18\times$--$4.75\times$ faster end-to-end inference.
\end{itemize}

\section{Related Work}

\subsection{Latent-Space Diffusion Models}
Latent Diffusion Models (LDMs) perform denoising in an autoencoder-derived latent space, substantially reducing computational and memory costs~\cite{ldm,vae}. Since the pioneering work on latent diffusion~\cite{ldm}, LDMs have advanced along several dimensions, including model architectures~\cite{dit,uvit,lumina2,sd3,pixartalpha,sdxl}, prediction paradigms~\cite{flow-matching,rectified-flow}, training and inference methods~\cite{dmd,flowgrpo,refl,repa,sra,dopsd,deepcache,teacache,cfg++,commondiff}, and distributed training and inference systems~\cite{xdit,diffusionpipe,distrifusion,sdxl,moviegen,gpu-varm-imagegen,z-reward}. Consequently, LDMs have become the de facto paradigm for large-scale text-to-image systems~\cite{hunyuanimage,qwen-image-2,z-image,flux-2,seedream4}, including the Seedream~\cite{seedream2,seedream3,seedream4} and FLUX~\cite{flux,flux-2} model families.

Despite their success, LDMs remain constrained by the autoencoder reconstruction bottleneck: aggressive latent compression can discard fine-grained details and impose an upper bound on sample fidelity~\cite{sd-vae,va-vae,dcae}. Training a large-scale autoencoder also requires substantial data, computational resources, and empirical tuning~\cite{qwen-image-vae,scale-tokenizer}. Moreover, the mismatch between the autoencoder's reconstruction objective and the diffusion model's generation objective can induce latent-space distribution shifts, such as smoothed textures and color distortions, that the diffusion process must compensate for~\cite{va-vae,repa-e}. Finally, the required VAE decoding stage introduces additional inference latency relative to direct pixel-space generation~\cite{l2p,zitpp,pixeldit}. These limitations motivate end-to-end alternatives such as pixel-space diffusion.

\subsection{Pixel-Space Diffusion Models}
Pixel-space diffusion predates latent-space diffusion, but early approaches were limited by computational cost and architectural bottlenecks~\cite{ddpm,adm,cdm}. With increased compute and advances in network design~\cite{pixnerd,dip,pixeldit,deco} and training objectives~\cite{jit,pixel-meanflow,pixelgen}, pixel-space methods have substantially narrowed the performance gap. On moderate-scale datasets such as ImageNet~\cite{imagenet}, they now achieve performance comparable to or better than latent-space methods. Unified Multimodal Models (UMMs) have also demonstrated the feasibility of pixel-space modeling for both visual understanding and generation at scale~\cite{tuna2,hidream-o1,sensenova-u1}. However, large-scale studies of pixel-space diffusion for generation-centric models remain scarce. In this work, we provide empirical insights and practical guidelines for training pixel-space diffusion models in the more challenging large-scale text-to-image setting.

\section{Large-Scale Pixel-Space Pre-training Converges More Slowly Than Latent-Space}
\label{sec:pre-training-compare}

First, a fundamental question remains unresolved: under controlled large-scale pre-training settings, how do pixel- and latent-space diffusion compare in terms of training efficiency and performance?

To answer this question, we conduct a controlled comparison using the same architecture, training data, computational resources, and training configuration as Z-Image~\cite{z-image}, varying only the prediction space.

\begin{figure}[h]
    \centering
    \includegraphics[width=1\linewidth]{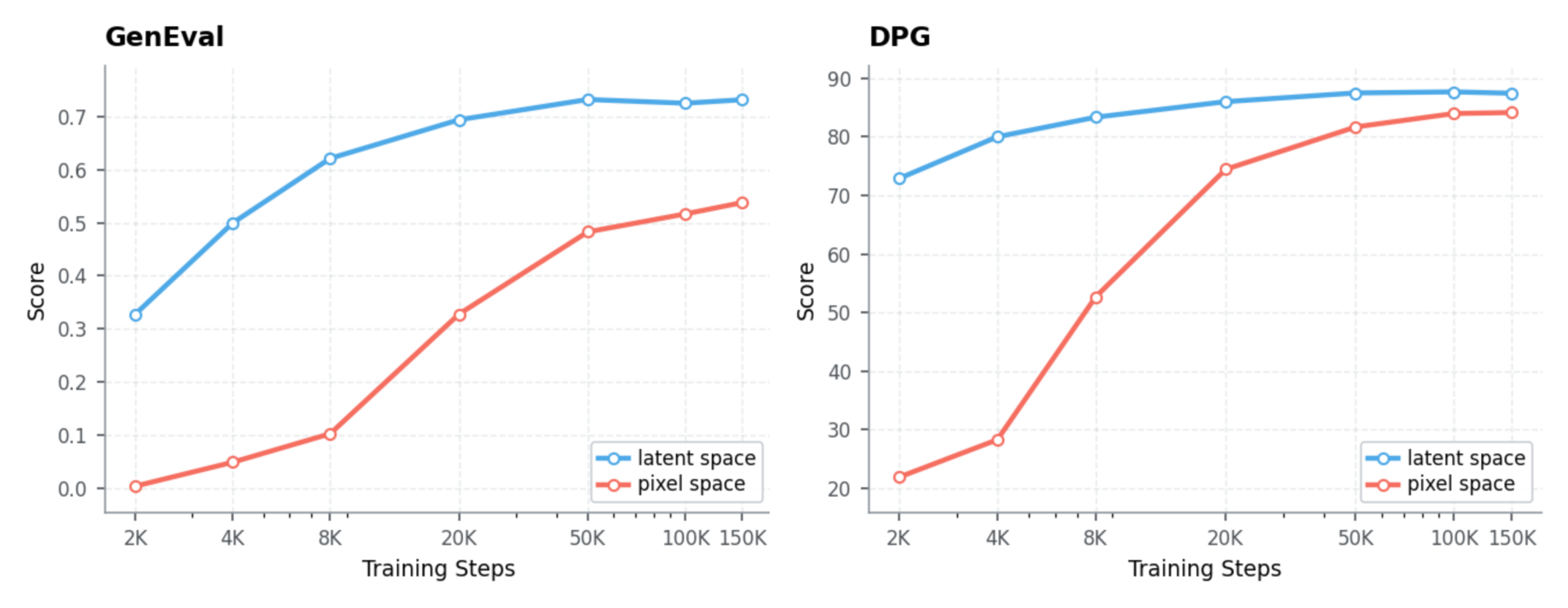}
    \caption{Evaluation curves over training steps. The model trained in the latent space consistently achieve superior results (measured by GenEval~\cite{geneval} and DPG~\cite{dpg}) and faster convergence compared to those trained in the pixel space.}
    \label{fig:latent-vs-pixel-bench}
\end{figure}

\begin{figure}[h]
    \centering
    \includegraphics[width=1\linewidth]{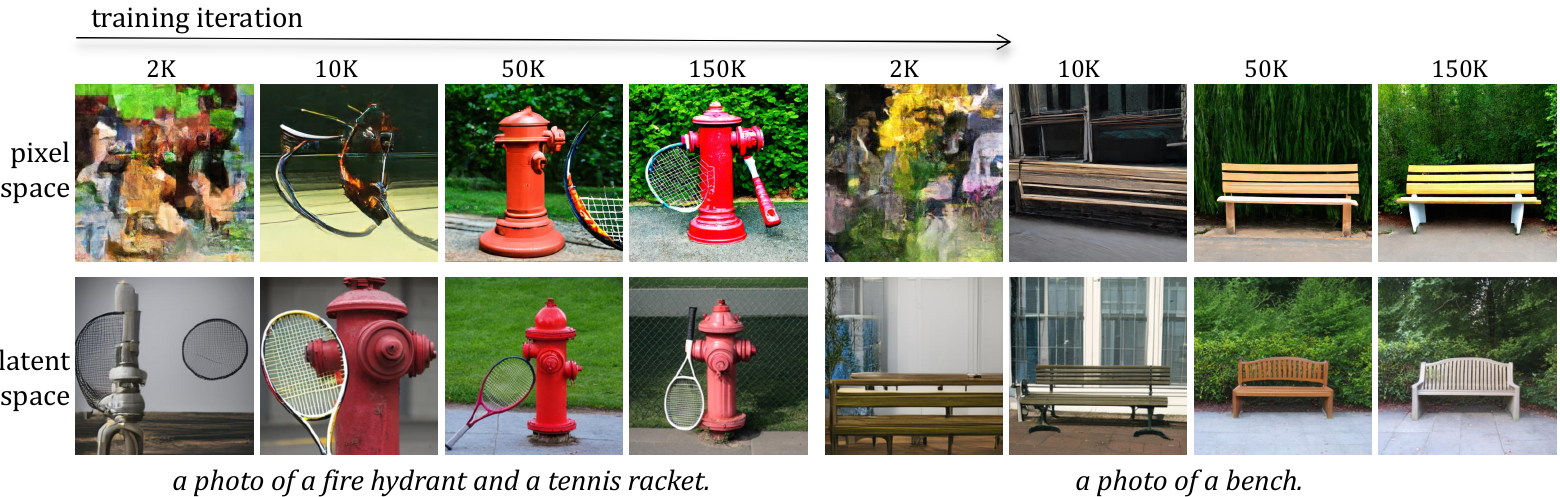}
    \caption{Qualitative samples over training iterations. The latent space model consistently demonstrates faster image structure learning and superior final image quality in pre-training progress.}
    \label{fig:latent-vs-pixel-vis}
\end{figure}

As shown in Figure~\ref{fig:latent-vs-pixel-bench}, latent-space pre-training consistently outperforms its pixel-space counterpart throughout training. The gap is particularly pronounced at the early stage: the latent model rapidly acquires basic object structure and text--image alignment, whereas the pixel model remains substantially behind. Although the pixel model continues to improve, it does not close the gap under the same compute budget. The qualitative results in Figure~\ref{fig:latent-vs-pixel-vis} show the same trend: latent-space samples exhibit recognizable global layouts and object identities much earlier, while pixel-space samples require more iterations to form coherent structures.


We hypothesize that this performance gap arises because the VAE serves not only as a computational compressor but also as a learned, compact visual representation~\cite{sd-vae,va-vae,dcae}. Trained on large-scale image data, the VAE maps raw pixels to a lower-dimensional, perceptually structured latent space that suppresses local redundancy and high-frequency variation while preserving information relevant to image semantics and appearance~\cite{ldm,vae}. This representation substantially simplifies the distribution that the diffusion model must learn. In contrast, a pixel-space model must discover these regularities directly from high-dimensional raw signals while simultaneously learning global image structure, modeling fine-grained local statistics, and performing denoising. This additional optimization burden makes learning from scratch considerably more challenging, even with the more than 20 billion image--text training pairs used in our experiments.

These results suggest that, despite avoiding the information loss introduced by latent compression, direct pixel-space supervision may not be the most effective formulation for large-scale pre-training because it converges more slowly and performs worse under a matched compute budget.

\takeaway{1}{
Under identical large-scale pre-training settings, latent-space diffusion learns image structure and text--image alignment substantially faster than pixel-space diffusion, suggesting that pixel-space training may be suboptimal for pre-training.
}

\section{A Carefully Designed Latent-to-Pixel Transition Offers a Better Trade-off}
\label{sec:l2p-ab}

As discussed in Section~\ref{sec:intro}, pixel-space diffusion is attractive for the final generator because it avoids the VAE reconstruction bottleneck and produces RGB images directly without an additional decoding stage. However, the results in Section~\ref{sec:pre-training-compare} show that direct pixel-space pre-training converges more slowly and achieves lower performance than latent-space pre-training under a matched compute budget. Together, these observations suggest that latent and pixel spaces are better suited to different stages of training. Large-scale pre-training primarily aims to acquire broad visual knowledge and text--image alignment efficiently, for which the compact latent representation provides a favorable trade-off between modest information loss and substantially easier optimization. Pixel-space supervision can then be introduced during post-training to obtain a final generator that operates directly in RGB space.

Motivated by  recent studies on latent-to-pixel adaptation~\cite{l2p,asymmetric-flow,crossflow}, we first pre-train a latent diffusion model and then adapt it to predict pixels directly. This strategy provides a strong and quick initialization for pixel-space training while retaining the quality and inference advantages of direct pixel generation. Using Z-Image~\cite{z-image} as the base model and following the general setup of L2P~\cite{l2p}, we  investigate the key design choices governing this transition. Detailed hyperparameter settings are provided in Appendix.

\subsection{Weight Initialization}
\label{subsec:weight-init}
\begin{figure}[h]
    \centering
    \includegraphics[width=1\linewidth]{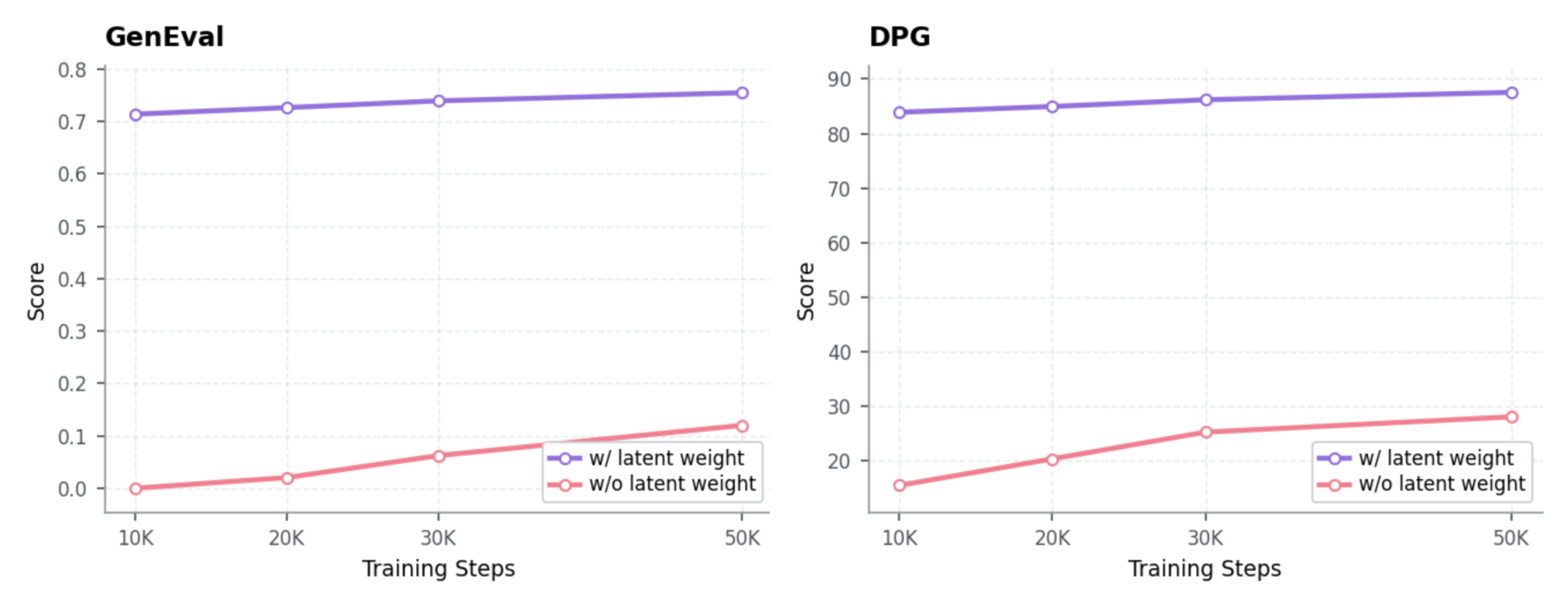}
   \caption{Pixel-space training that initialized from a pre-trained latent-space model yields remarkable faster convergence than training from scratch.}
    \label{fig:weight-bench}
\end{figure}

Given the substantially higher efficiency of latent-space pre-training, we first investigate whether the knowledge learned in latent space facilitates subsequent pixel-space optimization. Specifically, we compare two pixel-space models trained under identical settings: one initialized from a pre-trained latent-space model and the other trained from scratch. As shown in Figure~\ref{fig:weight-bench}, latent-space weight initialization yields consistently higher GenEval and DPG scores throughout training. In contrast, the model trained from scratch improves only gradually and remains far behind under the same training budget.

\begin{figure}[h]
    \centering
    \includegraphics[width=1\linewidth]{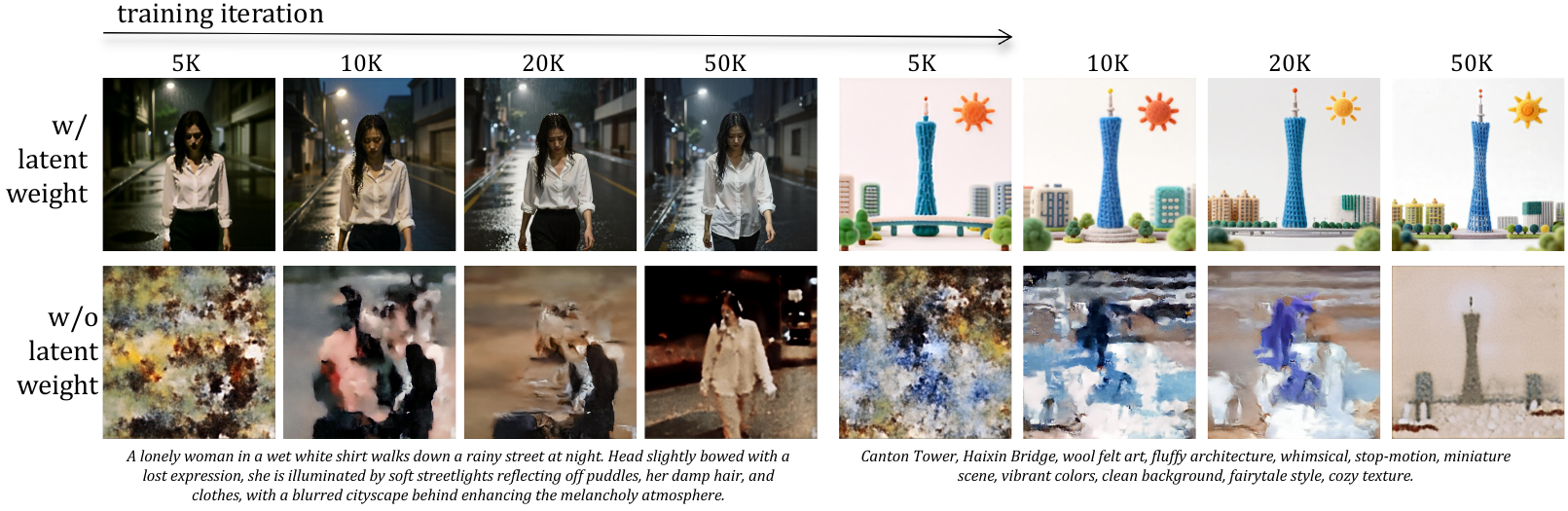}
    \caption{Pixel-Space trained with latent-weight initialization enables earlier formation of coherent image structures and improves final image quality compared with training from scratch.}
    \label{fig:weight-vis}
\end{figure}

The qualitative results in Figure~\ref{fig:weight-vis} show a similar pattern. Starting from latent-space weights enables the pixel-space model to form recognizable objects, coherent layouts, and stable text--image alignment much earlier, whereas training from scratch produces severely degraded and unstable samples for a substantially longer period.

\takeaway{2}{
The knowledge and generation capacity acquired in latent space remain highly transferable to pixel-space optimization, providing a strong initialization that accelerates convergence.
}

\subsection{Training Data}
\label{subsec:train-data}

\begin{figure}[h]
    \centering
    \includegraphics[width=1\linewidth]{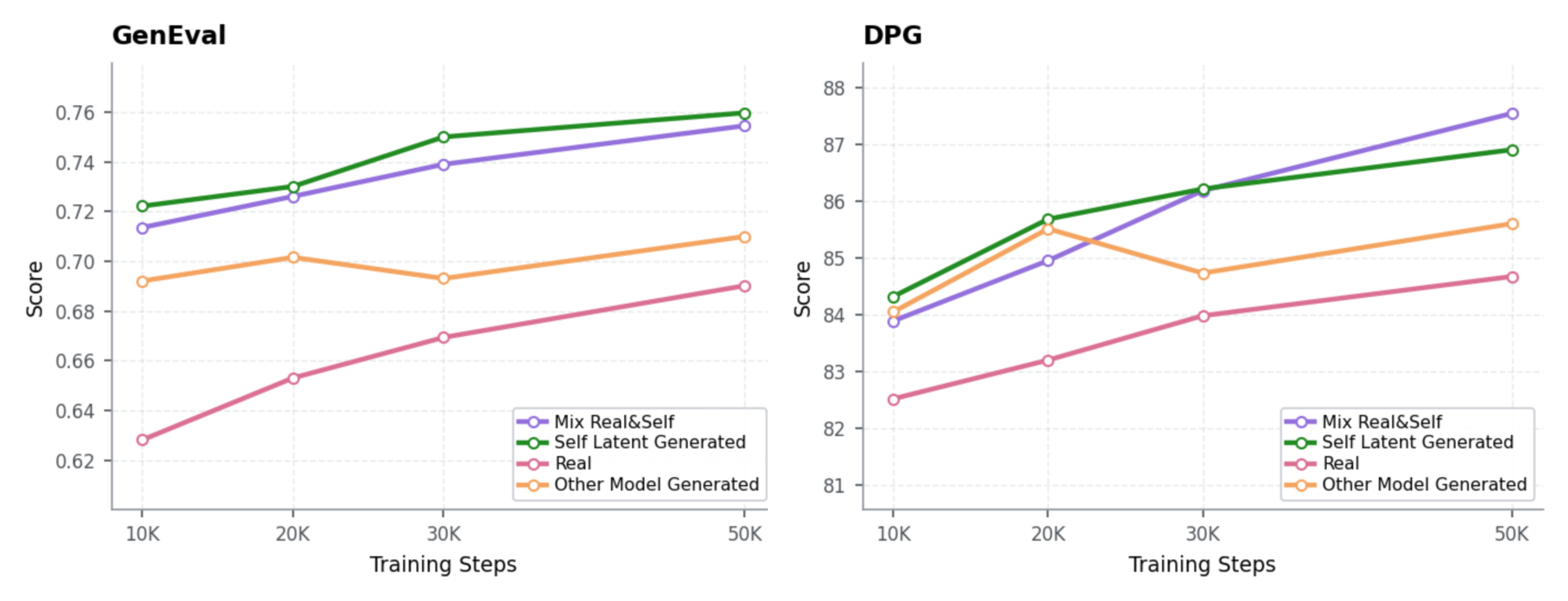}
    \caption{Effect of training data source. Source-latent-model-generated data enables substantially faster convergence than real data alone, while mixing real and self-generated data achieves the best overall trade-off.}
    \label{fig:train-data-bench}
\end{figure}

\begin{figure}[h]
    \centering
    \includegraphics[width=1\linewidth]{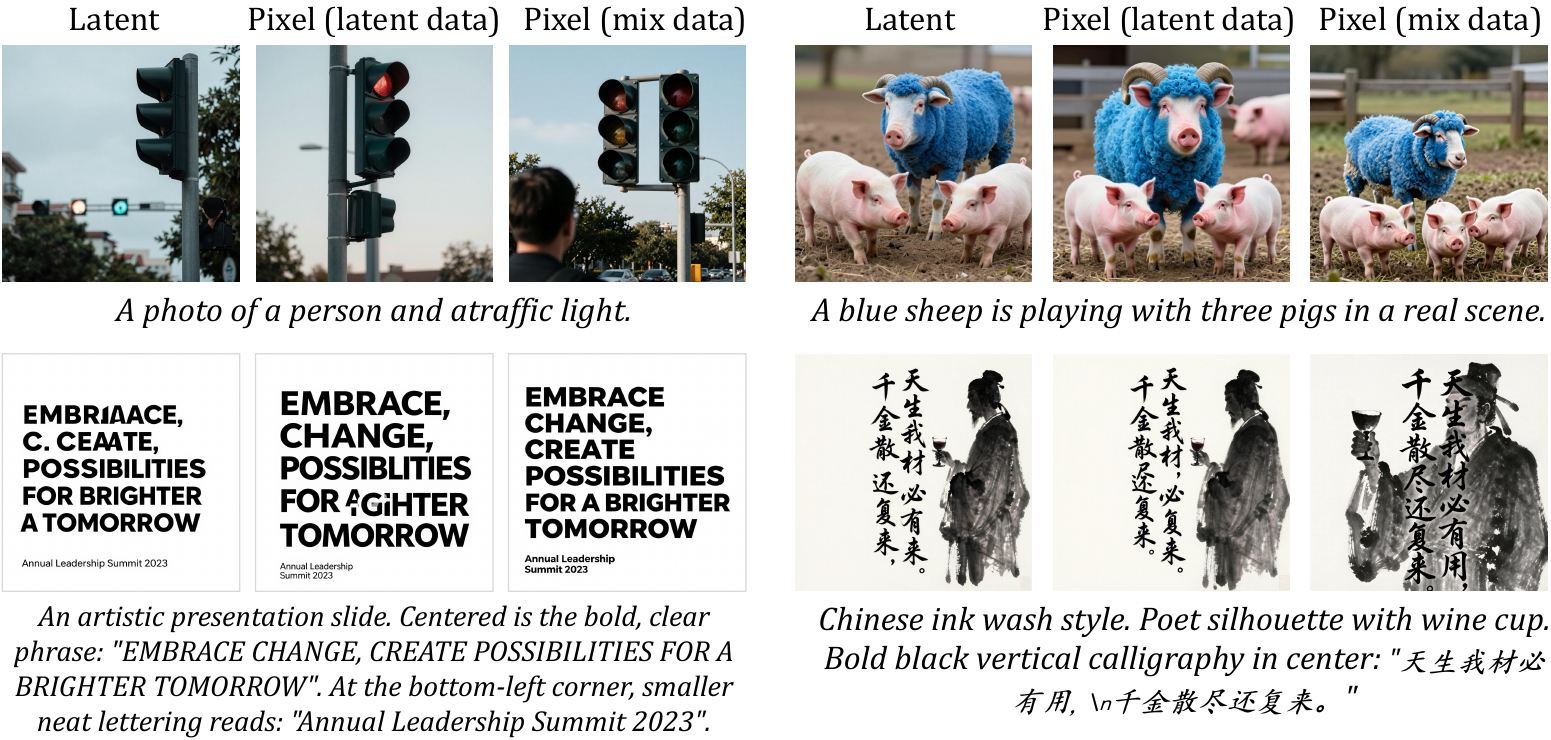}
    \caption{Qualitative comparison of training data choices. Training only on source latent-model generations can retain artifacts inherited from the source model. Mixing high-quality real images provides direct pixel-level supervision and improves these details.}
    \label{fig:train-data-vis}
\end{figure}

Pixel-space training enables direct supervision from high-quality real RGB images, allowing the final generator to move beyond the reconstruction bottleneck and information loss imposed by VAE compression. However, as shown in Figure~\ref{fig:train-data-bench}, training exclusively on real images converges substantially more slowly than the other settings.

In contrast, using samples generated by the same latent model that provides the initialization leads to much faster convergence. We hypothesize that these self-generated samples form a low-distribution-shift bridge between latent- and pixel-space training. Because they remain aligned with the source model's learned conditional distribution, they implicitly constrain the pixel-space model to preserve its existing generative priors while rapidly adapting to a new output representation. The weaker performance obtained with samples generated by a different model, FLUX2-klein-9B~\cite{flux-2}, supports this interpretation and suggests that alignment with the source-model distribution is important for effective transfer~\cite{l2p,dopsd,i-cm,zitpp}.

However, self-generated data inevitably inherits errors from the source latent model and its VAE decoder. As shown in Figure~\ref{fig:train-data-vis}, artifacts such as malformed typography persist when the pixel model is trained only on source-model samples. We therefore combine self-generated samples with high-quality real images. This mixture preserves the rapid and stable adaptation enabled by self-generated data, while real-image supervision recovers visual information lost through latent compression and corrects artifacts inherited from the source model, as shown in Figure~\ref{fig:train-data-vis}.

\takeaway{3}{
Self-generated data enables a smooth and rapid latent-to-pixel transition but inherits errors from the latent generator. Real-image supervision can correct these errors but converges slowly in isolation. Combining the two provides the best trade-off.
}

\subsection{Prediction Space}
\label{subsec:pred-space}

\begin{figure}[h]
    \centering
    \includegraphics[width=1\linewidth]{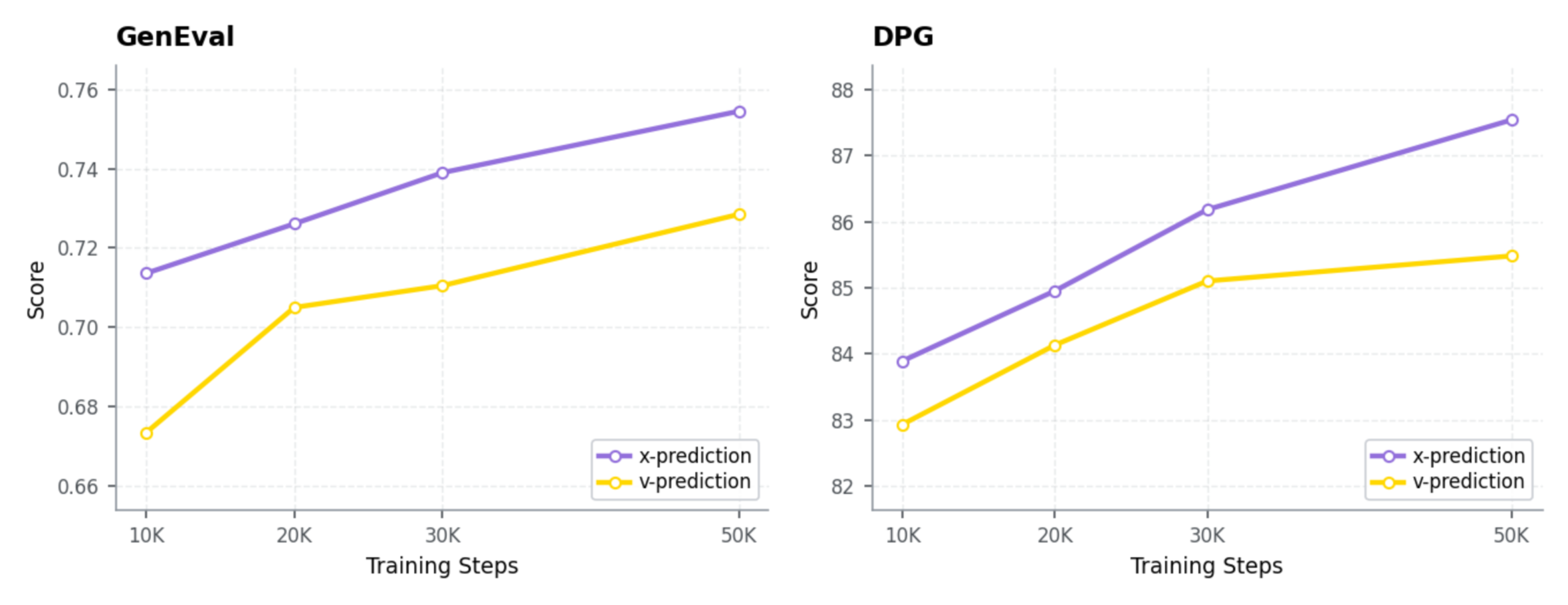}
   \caption{Training with  $x$-prediction achieves overall better performance and faster convergence than  $v$-prediction.}
    \label{fig:pred-space-bench}
\end{figure}

Prior studies such as JiT~\cite{jit} suggest that $x$-prediction is essential for stable pixel-space training, whereas the widely used $v$-prediction objective can lead to divergence. In our setting, however, the pixel-space model is initialized from a latent-space baseline trained with $v$-prediction. This setup raises a key design question for pixel-space post-training: should the model retain the original $v$-prediction objective or switch to $x$-prediction? As shown in Figure~\ref{fig:pred-space-bench}, the stable initialization provided by the latent-space weights enables both formulations to converge, avoiding the instability observed when training from scratch. Nevertheless, switching to $x$-prediction consistently yields better benchmark scores throughout training. These results indicate that latent-space initialization stabilizes the transition between prediction objectives, while $x$-prediction remains important for achieving the best pixel-space performance.

\takeaway{4}{
For pixel-space post-training, $x$-prediction consistently outperforms $v$-prediction, even when the model is initialized from a latent-space $v$-prediction model.
}

\begin{table*}[t]
\centering
\vspace{-1em}
\captionsetup{width=0.7\textwidth}
\caption{Decoder design comparison. Inference latency and GFLOPs are measured with decoder-only Inference for obtaining a $1024\times1024$ image on a single H800 GPU. The best and second-best results in each row are \textbf{bolded} and \underline{underlined}, respectively.}
 \vspace{-0.5em}
\begin{tabular}{c|cccc>{\color{gray}}c}
\toprule  
\diagbox{Metric}{Decoder} & JiT & Dip & PiT & Deco & VAE \\
\midrule  
GenEval $\uparrow$ & 0.7380 & \underline{0.7545} & \textbf{0.7697} & 0.7347 & - \\
DPG $\uparrow$ & 86.19 & \textbf{87.54} & \underline{86.36} & 86.10 & - \\
Params(M) $\downarrow$  & \textbf{2.95} & \underline{10.09} & 2249.82 & 315.92 & 49.55 \\
Latency (ms) $\downarrow$ & \textbf{0.07} & 6.02 & 74.96 & \underline{5.01} & 91.88\\
GFLOPs $\downarrow$ & \textbf{24.16} & \underline{834.03} & 19731.88 & 2587.05 & 10472.20 \\
\bottomrule 
\end{tabular}
\label{tab:decoder_comparison}
\end{table*}

\subsection{Decoder Architecture}
\label{subsec:decoder}

While DiT-style Transformer backbones have become standard for pixel-space diffusion~\cite{dit,sd3,lumina2}, the design of pixel decoder heads remains comparatively underexplored. Existing decoder designs have not been systematically compared in a controlled text-to-image setting. We therefore benchmark several representative designs, including JiT's linear head~\cite{jit}, DiP's lightweight convolutional U-Net~\cite{dip}, PixelDiT's Transformer-based head (PiT)~\cite{pixeldit}, Deco's frequency-decoupled head~\cite{deco}, and a conventional VAE for reference (FLUX-AE~\cite{flux}), while keeping the backbone and training setup fixed.

\begin{figure}[h]
    \centering
\includegraphics[width=1\linewidth]{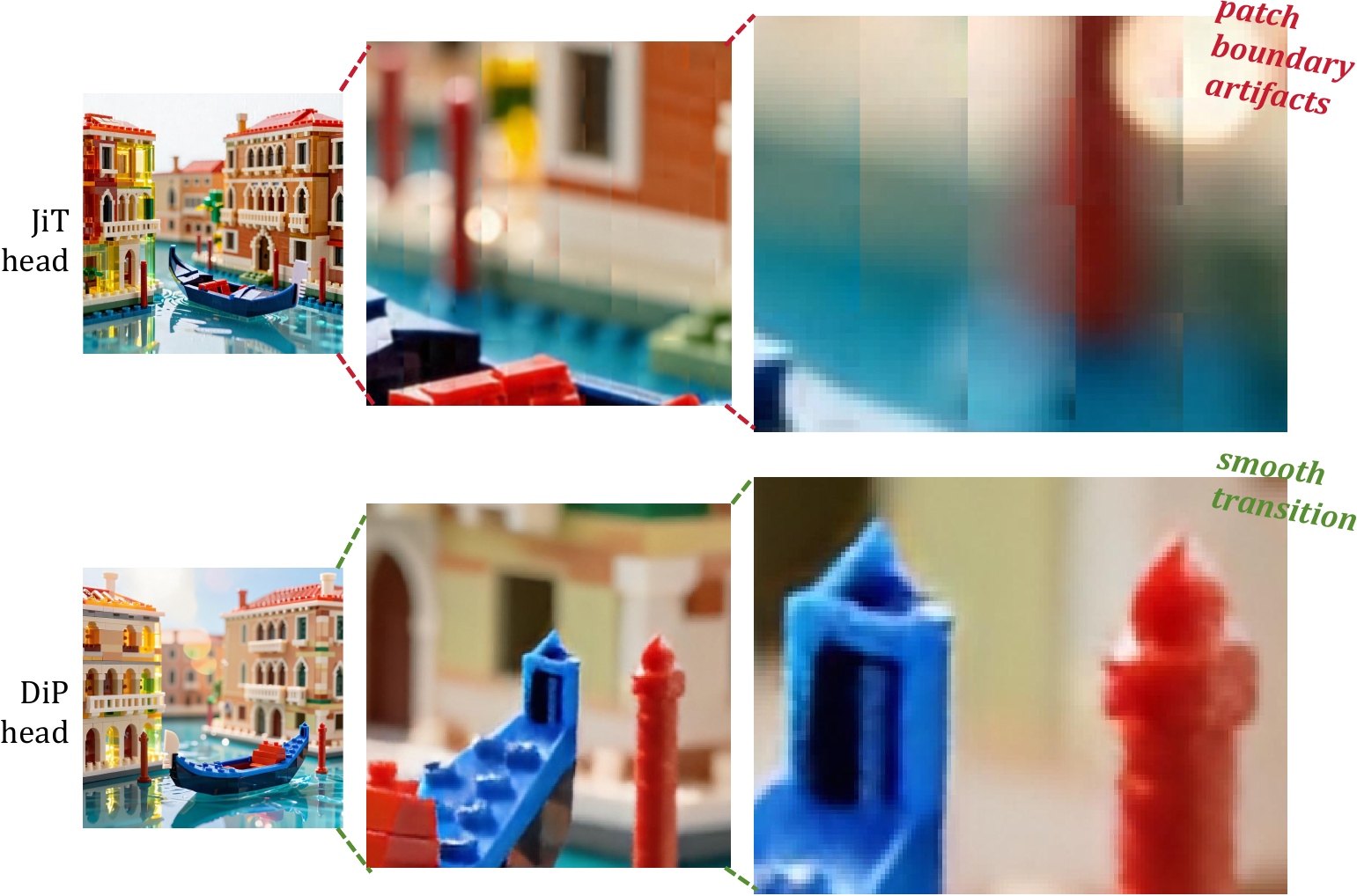}
    \caption{The JiT head exhibits visible grid-like artifacts at patch boundaries, whereas the DiP head produces smoother spatial transitions.}
    \label{fig:decoder-com-vis}
\end{figure}

The results in Table~\ref{tab:decoder_comparison} reveal a clear quality--efficiency trade-off. Although PiT achieves the highest GenEval score, its decoder alone contains 2.25B parameters, equivalent to 37.5\% of our 6B-parameter backbone, and requires 19,731 GFLOPs. This substantial decoder overhead makes PiT impractical for our setting. JiT is considerably more efficient but yields lower generation quality and visible grid-like artifacts at patch boundaries, as shown in Figure~\ref{fig:decoder-com-vis}.

DiP achieves the best DPG score and competitive GenEval performance with only 10.09M parameters and 834 GFLOPs, substantially reducing computation relative to PiT, Deco, and the VAE. It also produces smoother transitions across neighboring patches. We attribute this behavior to the local spatial inductive bias introduced by its lightweight convolutional decoder: convolutional locality and weight sharing allow neighboring patch features to interact before pixel synthesis. Such spatially shared processing may reduce discontinuities at patch boundaries without incurring the substantial cost of heavier decoder designs. Based on this quality--efficiency trade-off, we adopt DiP as our decoder head.

\takeaway{5}{
The DiP head provides the best quality--efficiency trade-off among the evaluated decoder designs and produces smoother transitions across patch boundaries.
}

\subsection{Noise Schedule}
\label{subsec:noise-schedule}

Transitioning from VAE latents to RGB pixels changes both the spatial resolution and the signal distribution on which the flow-matching path is defined. Consequently, directly reusing the latent-space noise schedule may expose the initialized model to a substantially different signal-to-noise ratio (SNR). Rather than redesigning the entire schedule, we follow previous studies~\cite{jit,sid,noise-schedule} and introduce a single noise-scale factor $\gamma$:
\begin{equation}
    \mathbf{x}_t = t\mathbf{x}_0 + (1-t)\gamma\boldsymbol{\epsilon},
    \qquad \boldsymbol{\epsilon}\sim\mathcal{N}(\mathbf{0},\mathbf{I}).
\end{equation}

We first derive a resolution-based reference for $\gamma$. Let $r$ denote the ratio between the spatial dimensions of an RGB image and its corresponding VAE latent. In our setting, $r=8$, meaning that the RGB representation has an $8\times$ higher spatial resolution along each dimension. If the pixel-space input is average-pooled by a factor of $r$, the variance of independent pixel noise decreases by $r^2$. The resulting SNR relationship is
\begin{equation}
    \operatorname{SNR}_{\mathrm{pixel}}^{\downarrow r}(t)
    \approx
    \frac{r^2}{\gamma^2}
    \operatorname{SNR}_{\mathrm{latent}}(t).
\end{equation}
Under the simplifying assumption that spatial resolution is the only difference between the two representations, matching their SNRs gives $\gamma=r=8$.

This derivation provides only a theoretical reference. A VAE encoder is not equivalent to average pooling, and RGB pixels and VAE latents also differ in their signal statistics and prediction targets. Resolution-based SNR matching therefore does not necessarily determine the optimal noise scale for latent-to-pixel adaptation. We accordingly evaluate $\gamma\in\{1,2,4,8\}$, using $\gamma=8$ as the theoretically motivated reference.

\begin{table}[t]
\centering
\caption{Ablation on noise scale $\gamma$.}
 \vspace{-0.5em}
\begin{tabular}{lcccc}
\toprule
\diagbox{Metric}{$\gamma$} & 1 & 2 & 4 & 8 \\
\midrule
GenEval $\uparrow$ & 0.6811 & \textbf{0.7545} & 0.7413 & 0.7316 \\
DPG $\uparrow$ & 84.03 & \textbf{87.54} & 86.01 & 85.64 \\
\bottomrule
\end{tabular}
\label{tab:noise-scale-ab-bench}
\end{table}

\begin{figure}[h]
    \centering
    \includegraphics[width=1\linewidth]{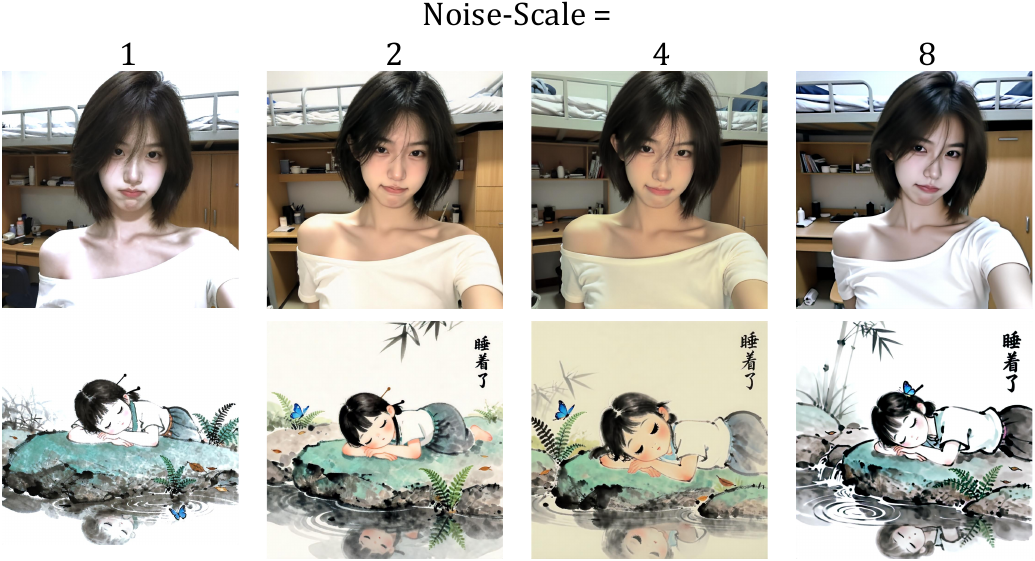}
    \caption{Qualitative comparison of different noise scales. Noise scales of $1$, $4$, and $8$ exhibit noticeable color shifts, whereas that of $2$ produces more balanced color rendition.}
    \label{fig:noise-scale-vis}
\end{figure}

As shown in Table~\ref{tab:noise-scale-ab-bench}, $\gamma=2$ achieves the best performance on both GenEval and DPG, whereas the resolution-derived value of $8$ is suboptimal. This discrepancy indicates that spatial resolution alone cannot fully characterize the distribution shift from VAE latents to RGB pixels and that empirical calibration remains necessary. The qualitative results in Figure~\ref{fig:noise-scale-vis} show the same trend: $\gamma=1$, $4$, and $8$ produce noticeable color shifts, while $\gamma=2$ yields more balanced color rendition for both realistic and stylized images. We therefore use $\gamma=2$ in subsequent experiments.

\takeaway{6}{
A resolution-based SNR analysis provides a useful reference for transferring the latent-space noise schedule, but the optimal noise scale must be empirically calibrated to account for the broader distribution shift between VAE latents and RGB pixels.
}

\section{Optimizing Efficiency with Larger-Patch Adaptation and Step Distillation}

\subsection{Larger Patch-Size Adaptation}
\label{subsec:larger-patch}

To further improve inference efficiency, one practical approach is to reduce the number of visual tokens by increasing the effective compression ratio. In latent-space diffusion, this can be achieved by increasing either the VAE compression ratio or the latent-space patch size. However, aggressive latent compression degrades reconstruction and generation quality~\cite{qwen-image-vae,dit,sd-vae}. Consequently, large-scale latent diffusion models typically use an effective $16\times$ spatial compression ratio, corresponding to $4{,}096$ tokens for $1024^2$ generation.

Pixel-space generation offers a potentially more favorable alternative. Because it is trained end-to-end with supervision in the original pixel space, recent work suggests that it can accommodate larger patch sizes without severe quality degradation~\cite{jit}. We therefore investigate larger patch sizes for large-scale text-to-image generation.

\begin{figure}[h]
    \centering
    \includegraphics[width=1\linewidth]{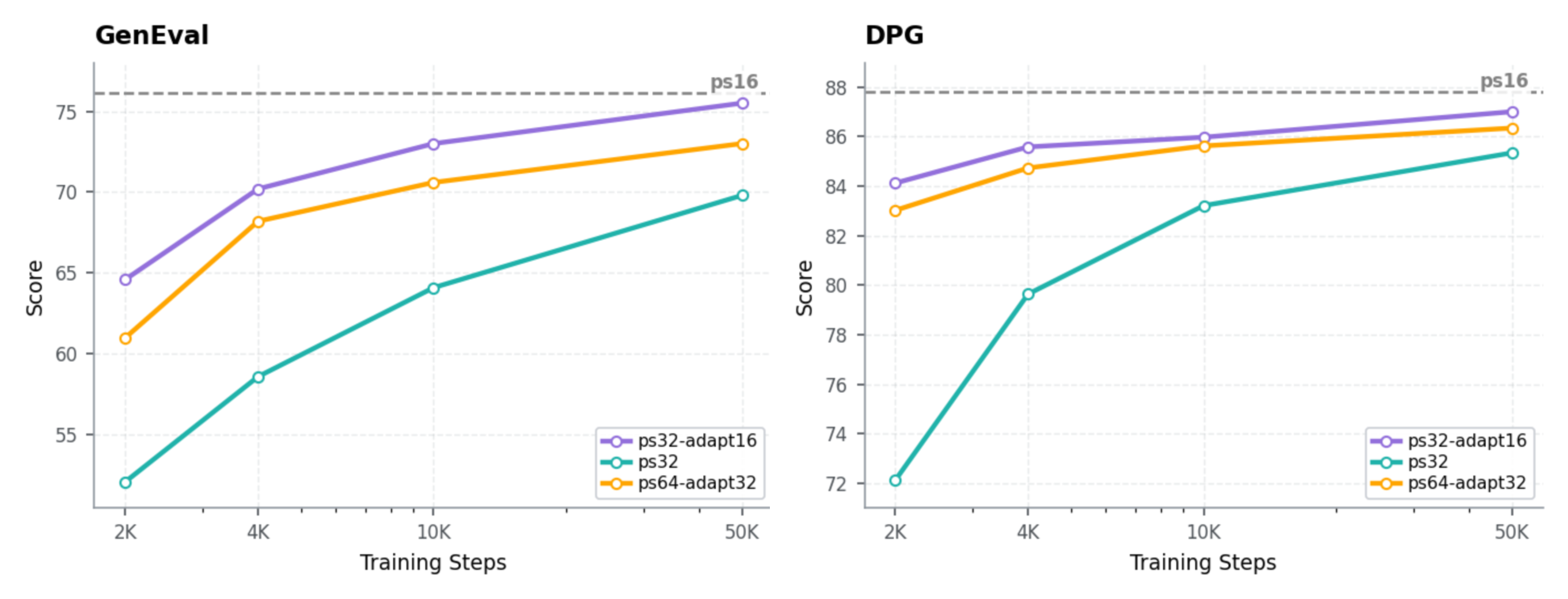}
    \caption{Progressive patch-size adaptation improves convergence and final performance.}
    \label{fig:patch-size-bench}
\end{figure}

The \texttt{ps16} baseline preserves the $64\times64$ spatial token grid of the $16\times$-compressed latent model. The latent-to-pixel transition therefore changes only the prediction space while retaining the spatial granularity at which the backbone learned its visual priors, leading to stable convergence and detailed synthesis.

\begin{figure}
    \centering
    \includegraphics[width=1\linewidth]{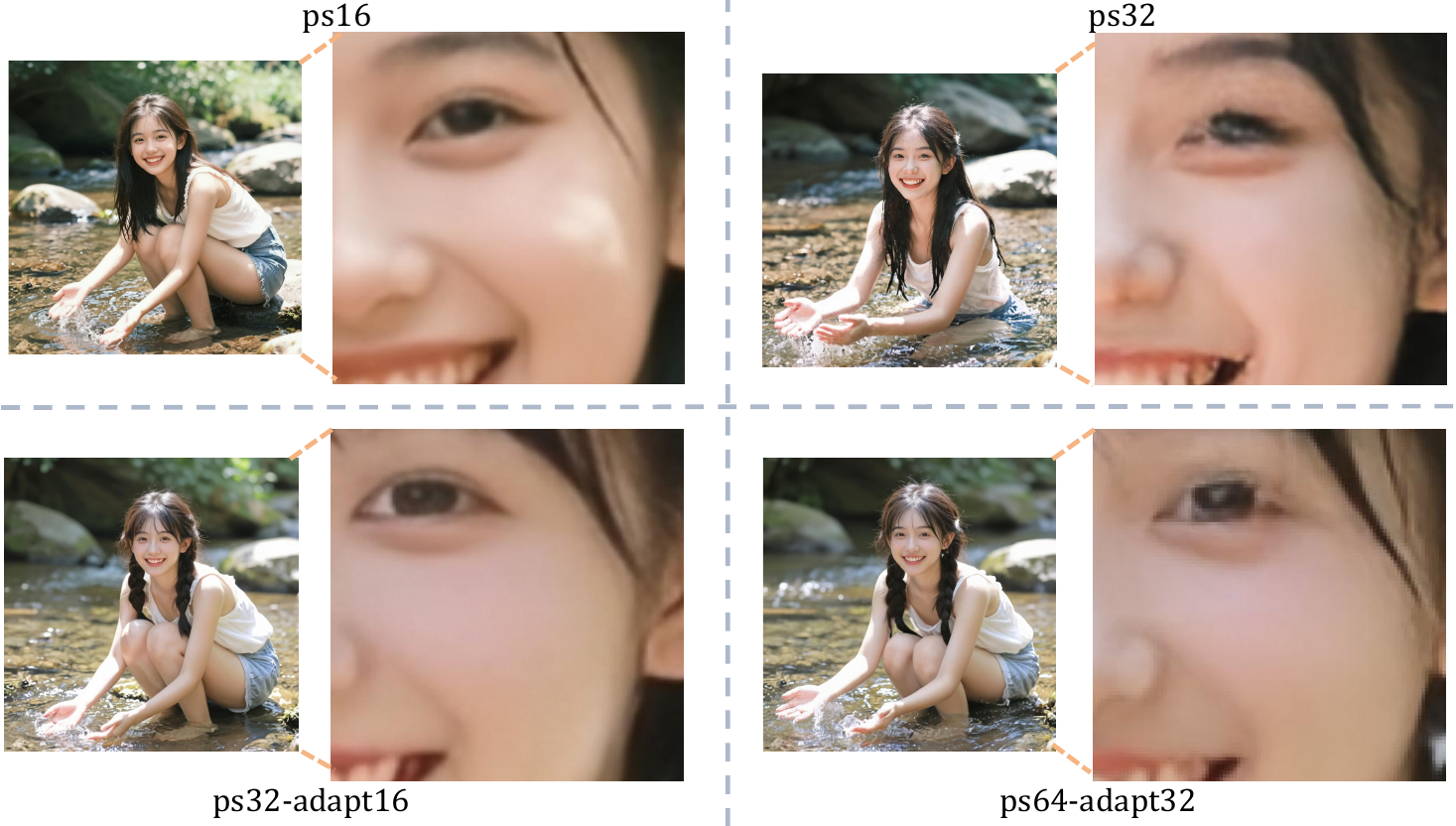}
    \caption{Qualitative comparison of patch-size adaptation. Direct \texttt{ps32} training introduces local artifacts. Adapting from \texttt{ps16} to \texttt{ps32} preserves these details, whereas further adaptation to \texttt{ps64} reintroduces artifacts.}
    \label{fig:patch-size-vis}
\end{figure}

By contrast, directly training \texttt{ps32} simultaneously changes the prediction space and coarsens the token grid. Each token must therefore model a larger spatial region, making local visual priors more difficult to transfer. Despite reducing the token count by $4\times$, direct \texttt{ps32} training converges slowly and produces visible local artifacts, as shown in Figures~\ref{fig:patch-size-bench} and~\ref{fig:patch-size-vis}.

We therefore first transfer the latent model to pixel space at \texttt{ps16} and then adapt it to \texttt{ps32}, a configuration denoted as \texttt{ps32-adapt16}. By decoupling the prediction-space transition from the change in spatial granularity, this strategy transfers stable pixel-space visual and local-detail priors to the more efficient configuration. \texttt{ps32-adapt16} converges substantially faster than direct \texttt{ps32} training, achieves performance comparable to \texttt{ps16} on both GenEval and DPG, and largely removes local artifacts while using only one quarter of the tokens. Extending the same procedure from \texttt{ps32} to \texttt{ps64} provides another $4\times$ token reduction but degrades local details and benchmark performance. Overall, \texttt{ps32-adapt16} provides the most favorable quality--efficiency trade-off.

\takeaway{7}{
Progressive patch-size adaptation reduces the number of visual tokens by fourfold with only a small quality gap. The degradation at more aggressive patch sizes highlights an important direction for future work: preserving fine-grained details under extreme token compression.
}

\begin{table*}[t]
\centering
\captionsetup{width=0.90\textwidth}
\caption{Performance comparison of pixel and latent models across benchmarks~\cite{geneval,dpg,oneig,longtext}. Images are generated using original benchmark prompts rather than Prompt-Enhanced (PE)~\cite{prompt-enhancer} variants. Inference latency is measured for a $1024\times1024$ image on a single H800 GPU without any system-level optimizations.}
\vspace{-0.5em}
\label{tab:sys_comparison}
\begin{tabular}{lllcccc} 
\toprule
Model & NFE & Latency (s) $\downarrow$ & GenEval $\uparrow$ & DPG $\uparrow$ & OneIG $\uparrow$ & LongText $\uparrow$ \\
\midrule
Z-Image (latent-space) & 100 & 20.12 & 0.7510 & 86.91 & \textbf{0.566} & \textbf{0.9332}  \\
L2P (pixel-space)& 100 & 18.26 & 0.7612 & 86.00 & 0.499 & 0.7798 \\
\textbf{Ours (pixel-space)} & 100 &
\textbf{4.56}~\textcolor{green!50!black}{(4.41$\times$)} &
\textbf{0.7644} & \textbf{87.60} & 0.564 & 0.9252 \\ 
\arrayrulecolor{gray}\midrule\arrayrulecolor{black}
Z-Image-Turbo (latent-space)& 4 & 0.95 & 0.7625 & 85.31 & \textbf{0.520} & 0.8639   \\
\textbf{Ours-Turbo (pixel-space)} & 4 &
\textbf{0.20}~\textcolor{green!50!black}{(4.75$\times$)} &
\textbf{0.7698} & \textbf{86.85} & 0.512 & \textbf{0.8732} \\ 
\midrule
FLUX2-klein-Base (latent-space) & 100 & 20.25 & 0.8027 & 84.89 & 0.541 & 0.5456   \\
 AsymFlow (pixel-space)& 100 & 21.42 & \textbf{0.8181} & 86.70 & 0.508 & 0.4498 \\
\textbf{Ours (pixel-space)} & 100 &
\textbf{6.38}~\textcolor{green!50!black}{(3.18$\times$)} &
0.8096 & \textbf{86.88} & \textbf{0.554} & \textbf{0.6632}    \\ 
\arrayrulecolor{gray}\midrule\arrayrulecolor{black}
FLUX2-klein (latent-space) & 4 & 0.92 & 0.8142 & 84.36 & \textbf{0.537} & 0.4763    \\
\textbf{ Ours-Turbo (pixel-space)} & 4 &
\textbf{0.28}~\textcolor{green!50!black}{(3.29$\times$)} &
\textbf{0.8185} & \textbf{86.00} & 0.530 & \textbf{0.6105} \\
\bottomrule
\end{tabular}
\end{table*}

\subsection{Pixel-Space Step-Distillation}
\label{subsec:step-distillation}

Step distillation complements patch-size adaptation by reducing the number of function evaluations (NFEs) required for sampling~\cite{dmd,dmdr,ddmd,hypersd,tdm,dmd2,add}. In latent-space diffusion, however, its end-to-end acceleration becomes increasingly limited in the ultra-few-step regime. As denoising becomes cheaper, the fixed cost of a single VAE decoding pass accounts for a larger fraction of the total inference latency~\cite{zitpp}.

\begin{figure}[h]
    \centering
    \includegraphics[width=1\linewidth]{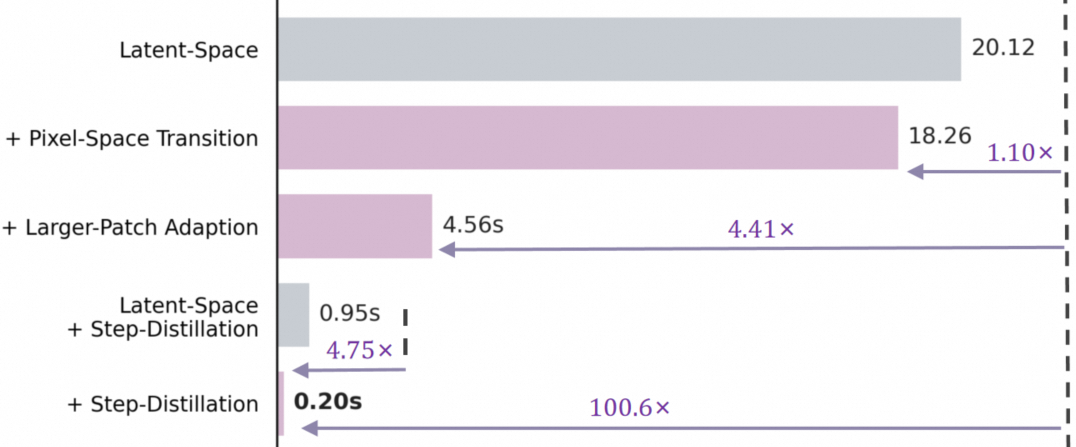}
    \caption{End-to-end inference latency (S) per $1024\times1024$ image on a single NVIDIA H100 GPU without any system-level optimizations such as FlashAttention~\cite{flashattention}.}
    \label{fig:inference-speed}
\end{figure}

Pixel-space generation removes this bottleneck by producing RGB images directly. We therefore apply Decoupled-DMD~\cite{ddmd} and DMDR~\cite{dmdr} to pixel-space step distillation, allowing reductions in NFE to translate more directly into end-to-end speedups. As shown in Figure~\ref{fig:inference-speed}, the pixel-space transition alone provides a modest $1.10\times$ speedup, while larger-patch adaptation reduces latency to $4.56$\,s ($4.41\times$). After step distillation, the pixel-space model requires only $0.20$\,s per image, compared with $0.95$\,s for its latent-space distilled counterpart, yielding a $4.75\times$ reduction in latency. Combined with larger-patch adaptation, pixel-space step distillation achieves a $100.6\times$ end-to-end speedup over the original latent-space pipeline while maintaining performance, as shown in Table~\ref{tab:sys_comparison}.

\section{System-Level Comparison}
\label{sec:system-comparison}

Having established the key design choices for pixel-space training, we scale up the mixed training data and extend training to obtain our final models. Table~\ref{tab:sys_comparison} evaluates our recipe on two model families. At 100 NFE, we compare against both the corresponding latent-space models and prior pixel-space adaptation methods: L2P~\cite{l2p} for Z-Image and AsymFlow~\cite{asymmetric-flow} for FLUX2-klein. At 4 NFE, we compare our distilled models with the latent-space official counterparts.

For Z-Image, our pixel-space models maintain comparable or better overall performance than their latent-space counterparts while achieving $4.41\times$ and $4.75\times$ speedups at 100 and 4 NFE. They also outperform L2P across all four benchmarks with substantially lower latency. Although our empirical study primarily uses Z-Image, the same recipe transfers effectively to FLUX2-klein~\cite{flux-2}, delivering $3.18\times$--$3.29\times$ speedups while achieves better results than both the latent-space baselines and AsymFlow across majority of benchmarks. 

These results suggest that the findings generalize beyond a single model family and provide a practical recipe for training efficient, high-performing pixel-space models and show that our recipe outperforms prior latent-to-pixel methods on most benchmarks, while maintaining competitive performance against well-established latent-space counterparts  across two model families, while offering substantially faster inference speed.

\section{Conclusion}

In this work, we systematically study large-scale pixel-space text-to-image diffusion and show that direct pixel-space pre-training converges substantially more slowly than latent-space pre-training. This finding motivates a latent-to-pixel strategy that acquires generative priors efficiently in latent space before adapting the model to pixel space during post-training. By examining the key design choices underlying this transition, we derive a practical recipe that outperforms prior latent-to-pixel methods on most benchmarks across two model families, while maintaining competitive performance against latent-space counterparts and delivering $3.18\times$--$4.75\times$ end-to-end inference speedups. These findings provide practical guidance for scalable, high-quality, and efficient pixel-space diffusion.

\section{Acknowledgments}
We thank Qilong Wu, Xin Jin, Zhen Li, Shilin Zhou, Zechao Zhan, and Ruikai Zhou for their helpful discussions and suggestions throughout this project.
 
{
    \small
    \bibliographystyle{ieeenat_fullname}
    \bibliography{main}
}

\clearpage
\appendix
\twocolumn[
  \begin{@twocolumnfalse}
    \section*{\centering Appendix}
  \end{@twocolumnfalse}
]

\providecommand{\tbd}[1]{\textcolor{red}{[TBD: #1]}}

\section{Additional Implementation Details}
\label{app:implementation-details}

In this section, we provide additional implementation details for the experiments presented in the main paper. The configurations used in our component-wise studies are summarized in Tables~\ref{tab:detailed_design_1} and~\ref{tab:detailed_design_2}. Unless otherwise specified, we keep the backbone architecture, text-conditioning modules, data preprocessing, optimization strategy, and evaluation protocol fixed when comparing different configurations.

\begin{table*}[t]
    \centering
    \caption{\textbf{Hyperparameter settings for component-wise analysis (I).}
    Bold entries denote the configurations varied in the corresponding section.}
    \vspace{-0.5em}
    \label{tab:detailed_design_1}

    \begingroup
    \setlength{\tabcolsep}{6pt}
    \renewcommand{\arraystretch}{1.15}
    \begin{tabular}{
        @{}
        >{\raggedright\arraybackslash}m{0.18\textwidth}
        >{\centering\arraybackslash}m{0.245\textwidth}
        >{\centering\arraybackslash}m{0.245\textwidth}
        >{\centering\arraybackslash}m{0.245\textwidth}
        @{}
    }
        \toprule
        & \textbf{Section~\ref{subsec:weight-init}}
        & \textbf{Section~\ref{subsec:train-data}}
        & \textbf{Section~\ref{subsec:pred-space}} \\
        \midrule

        Weight initialization
        & \shortstack{\textbf{From scratch}/\\
                      \textbf{Latent-pretrained}}
        & Latent-pretrained
        & Latent-pretrained \\

        Training data
        & Self-generated + Real
        & \shortstack{\textbf{Real images}/\\
                      \textbf{Self-generated samples}/\\
                      \textbf{Other-model samples}/\\
                      \textbf{Self-generated + Real}}
        & Self-generated + Real \\

        Prediction target
        & $x$-prediction
        & $x$-prediction
        & \shortstack{\textbf{$v$-prediction}/\\
                      \textbf{$x$-prediction}} \\

        Decoder architecture
        & DiP
        & DiP
        & DiP \\

        Noise scale
        & $\gamma=2$
        & $\gamma=2$
        & $\gamma=2$ \\

        Patch size
        & \texttt{ps16}
        & \texttt{ps16}
        & \texttt{ps16} \\

        \bottomrule
    \end{tabular}
    \endgroup
\end{table*}

\begin{table*}[t]
    \centering
    \caption{\textbf{Hyperparameter settings for component-wise analysis (II).}
    Bold entries denote the configurations varied in the corresponding section.}
    \label{tab:detailed_design_2}
\vspace{-0.5em}
    \begingroup
    \setlength{\tabcolsep}{6pt}
    \renewcommand{\arraystretch}{1.15}
    \begin{tabular}{
        @{}
        >{\raggedright\arraybackslash}m{0.18\textwidth}
        >{\centering\arraybackslash}m{0.245\textwidth}
        >{\centering\arraybackslash}m{0.245\textwidth}
        >{\centering\arraybackslash}m{0.245\textwidth}
        @{}
    }
        \toprule
        & \textbf{Section~\ref{subsec:decoder}}
        & \textbf{Section~\ref{subsec:noise-schedule}}
        & \textbf{Section~\ref{subsec:larger-patch}} \\
        \midrule

        Weight initialization
        & Latent-pretrained
        & Latent-pretrained
        & Latent-pretrained \\

        Training data
        & Self-generated + Real
        & Self-generated + Real
        & Self-generated + Real \\

        Prediction target
        & $x$-prediction
        & $x$-prediction
        & $x$-prediction \\

        Decoder architecture
        & \shortstack{\textbf{JiT}/
                      \textbf{DiP}/ \\
                      \textbf{PiT}/
                      \textbf{Deco}/ \\
                      \textbf{VAE}}
        & DiP
        & DiP \\

        Noise scale
        & $\gamma=2$
        & \textbf{$\gamma\in\{1,2,4,8\}$}
        & $\gamma=2$ \\

        Patch size
        & \texttt{ps16}
        & \texttt{ps16}
        & \shortstack{\textbf{\texttt{ps16}}/
                      \textbf{\texttt{ps32}}/\\
                      \textbf{\texttt{ps32-adapt16}}/\\
                      \textbf{\texttt{ps64-adapt32}}} \\

        \bottomrule
    \end{tabular}
    \endgroup
\end{table*}

\subsection{Large-Scale Pixel Versus Latent-Space pre-training}
For the controlled comparison in Sec.~\ref{sec:pre-training-compare}, we instantiate both models using the same Z-Image Transformer backbone~\cite{z-image} and text-conditioning stack (Pre-trained Qwen-3-4B~\cite{qwen3}). The latent-space model operates on features produced by the Flux-AE as used in Z-Image, which has a spatial downsampling factor of $8$. Together with a latent patch size of $2$, this corresponds to an effective spatial compression ratio of $16$ and a $64\times64$ token grid for $1024^2$ image generation. The pixel-space model uses \texttt{ps16}, producing the same $64\times64$ token grid directly from RGB images. Consequently, the two models have identical Transformer sequence lengths, allowing the comparison to primarily isolate the effect of the prediction space rather than differences in backbone computation.

Both models are trained on the same corpus of more than 20B image--text pairs following the original Z-Image pre-training recipe, including its progressive resolution schedule from $256^2$ to $512^2$, batch size, learning rate, optimizer, etc., while varying only the prediction space.

\subsection{Latent-to-Pixel Transition}
For the experiments in Sec.~\ref{sec:l2p-ab}, we initialize the pixel-space model from a Z-Image latent-space checkpoint.  During pixel-space post-training, we optimize the full model, with a global batch size of 128, a learning rate of 5e-5 and trained directly on $1K$ resolution following L2P~\cite{l2p}. Unless explicitly ablated, we use latent-pretrained initialization, mixed self-generated and real data, $x$-prediction, the DiP decoder, a noise scale of $\gamma=2$, and \texttt{ps16}.

\noindent\textbf{Weight initialization.}
For the initialization study in Sec.~\ref{subsec:weight-init}, the latent-initialized model loads the Transformer and conditioning weights from the pre-trained latent model, while its pixel-specific input and output modules are initialized identically to those of the from-scratch baseline. For the latter, the complete model is initialized using the initialization scheme of Z-Image. The two configurations use the same architecture, mixed training data, optimizer, learning-rate schedule, and number of updates.

\noindent\textbf{Training data.}
For the data study in Sec.~\ref{subsec:train-data}, we construct four training configurations: real images only, self-generated samples only, samples generated by another model, and a mixture of self-generated and real images. The real-image subset contains the latest Supervised-Fine-tuning (SFT) data of the Z-Image. Self-generated samples are produced by the same latent-space Z-Image checkpoint used for weight initialization with the default sampling settings, ensuring that the synthetic distribution remains close to the source model. 
For the external-model setting, we use FLUX2-klein-9B~\cite{flux-2} to generate images from the same prompt pool. Synthetic samples are generated offline  and paired with their original prompts. The mixed-data configuration samples self-generated and real images at a ratio of 1:1.

\noindent\textbf{Prediction space.}
For Sec.~\ref{subsec:pred-space}, we compare $v$-prediction and $x$-prediction under an otherwise identical latent-to-pixel setup. Both variants use the same flow-matching path, timestep sampler, noise scale, initialized backbone, and pixel decoder. The $x$-prediction model directly estimates the clean RGB target $\mathbf{x}_0$~\cite{jit}, whereas the $v$-prediction model estimates the velocity associated with the interpolation path~\cite{rectified-flow}. Timesteps are sampled from a logit-normal distribution~\cite{sd3}. Following JiT~\cite{jit}, both parameterizations are optimized using the same velocity-space objective. For $x$-prediction, the predicted clean image is converted into the corresponding velocity before computing the loss and performing ODE sampling. This allows both variants to share the same training weighting, numerical solver, and sampling schedule. During inference, the $x$-prediction output is converted to the corresponding velocity field. This conversion allows both variants to use the same numerical solver and sampling schedule.

\noindent\textbf{Decoder architecture.}
For the decoder comparison in Sec.~\ref{subsec:decoder}, all variants use the same latent-initialized backbone, training data, $x$-prediction objective, noise scale, and \texttt{ps16} tokenization. Only the module that maps the final Transformer features to RGB pixels is changed. The JiT head uses a linear projection to reconstruct the pixels associated with each visual token. The DiP head design remains consistent with L2P~\cite{l2p} which uses a lightweight convolutional U-Net. The PiT head consists of 4 additional Transformer blocks following~\cite{pixeldit}. For Deco, we use frequency decomposition and decoder configuration for training. The VAE reference uses the FLUX-AE used in Z-Image.

All decoder-specific modules are initialized from scratch, while the shared Transformer backbone is initialized from the same latent checkpoint. Parameter counts and GFLOPs are measured at a resolution of $1024^2$ with a batch size of one.

\noindent\textbf{Noise schedule.}
For Sec.~\ref{subsec:noise-schedule}, RGB images are normalized to [-1,1] before noise is applied. We evaluate $\gamma\in\{1,2,4,8\}$ in
\begin{equation}
    \mathbf{x}_t
    =
    t\mathbf{x}_0
    +
    (1-t)\gamma\boldsymbol{\epsilon},
    \qquad
    \boldsymbol{\epsilon}\sim\mathcal{N}(\mathbf{0},\mathbf{I}).
\end{equation}
The timestep distribution, prediction target, loss weighting, initialization, and optimization schedule are identical across all four settings. During sampling, the same value of $\gamma$ used for training is incorporated into noise initialization. We use $\gamma=2$ in all subsequent experiments based on its performance on GenEval and DPG.

\paragraph{Larger patch-size adaptation.}
For Sec.~\ref{subsec:larger-patch}, \texttt{ps16}, \texttt{ps32}, and \texttt{ps64} produce token grids of $64\times64$, $32\times32$, and $16\times16$, respectively, at $1024^2$ resolution. Direct \texttt{ps32} training initializes the shared Transformer from the latent-space checkpoint while simultaneously replacing the latent input representation with $32\times32$ RGB patches. In contrast, \texttt{ps32-adapt16} is initialized from the converged \texttt{ps16} pixel-space checkpoint, thereby decoupling the latent-to-pixel transition from the subsequent reduction in token resolution.

For progressive patch-size adaptation, the Transformer and decoder architectures remain unchanged, and only the input dimension of the first projection MLP must be expanded. When doubling the patch size from $p$ to $2p$, each new patch comprises four non-overlapping $p\times p$ sub-patches. Given the previous projection weight $\mathbf{W}$, we initialize the expanded weight as:
\begin{equation}
    \mathbf{W}'=\frac{1}{2}
    \left[\mathbf{W},\mathbf{W},\mathbf{W},\mathbf{W}\right],
\end{equation}
where the weights are replicated along the input dimension and the bias is copied directly. The factor of $1/2$ compensates for the four-fold increase in fan-in and approximately preserves the activation variance at initialization. All remaining Transformer and decoder parameters are transferred without modification. The same strategy is used when initializing \texttt{ps64-adapt32} from the converged \texttt{ps32-adapt16} checkpoint.

During adaptation, we apply a linear learning-rate warm-up of $5{,}000$ steps to the transferred Transformer and decoder parameters. The expanded input-projection MLP is exempt from this warm-up and is optimized with its target learning rate from the first update, allowing it to adapt rapidly to the new patch representation. We train \texttt{ps16}, \texttt{ps32-adapt16}, and \texttt{ps64-adapt32} for each stage, while keeping the training data, $x$-prediction objective, DiP decoder, and noise scale fixed.

\paragraph{Pixel-space step distillation.}
For Sec.~\ref{subsec:step-distillation}, we use the converged ps32-adapt16 pixel-space model for DMD initialization. We apply Decoupled-DMD~\cite{ddmd} in distillation stage 1 and DMDR~\cite{dmdr} with Z-Reward~\cite{z-reward} in stage 2. The corresponding loss coefficients, fake-sample update frequency, and timestep-sampling distribution are set following the DMD2 and Decoupled-DMD papers~\cite{dmd2,ddmd}.

The distilled model uses four function evaluations with an ODE sampler and does not require classifier-free guidance (CFG)~\cite{cfg}. For a controlled efficiency comparison, the latent-space distilled baseline uses the same number of function evaluations for testing.

\subsection{System-Level Comparison}
For the final comparison in Sec.~\ref{sec:system-comparison}, we scale up the mixed training set and extend the pixel-space post-training iteration. We apply the resulting recipe---latent initialization, mixed self-generated and real data, $x$-prediction, the DiP decoder, $\gamma=2$, and progressive patch-size adaptation---to both Z-Image and FLUX2-klein. For FLUX2-klein, we initialize from FLUX2-klein-9B-Base and use a synthetic-to-real data ratio of $1{:}1$. We otherwise retain the same training recipe, introducing only the architecture-specific modifications required by FLUX2-klein. As for Z-Image-L2P and FLUX2-klein-AsymFlow, we use the inference setting and checkpoint provided in the official open-source repository for benchmark evaluation.

\end{document}